\documentclass[a4paper, 10pt, conference]{ieeeconf}      
\usepackage{FG2017}
\usepackage{times}
\usepackage{epsfig}
\usepackage{graphicx}
\usepackage{amsmath}
\usepackage{amssymb}
\usepackage{amssymb}
\usepackage{amsfonts}
\usepackage{cite}
\usepackage{url}

\usepackage{caption} 
\usepackage{subcaption}
\usepackage{tabularx,ragged2e}
\usepackage{booktabs,calc}
\usepackage{multirow}
\usepackage{hhline}
\usepackage{array}
\usepackage{float}
\restylefloat{table}
\usepackage{makecell}
\usepackage{color}
\usepackage{authblk}
\newcolumntype{L}[1]{>{\raggedright\let\newline\\\arraybackslash\hspace{0pt}}m{#1}}
\newcolumntype{C}[1]{>{\centering\let\newline\\\arraybackslash\hspace{0pt}}m{#1}}
\newcolumntype{R}[1]{>{\raggedleft\let\newline\\\arraybackslash\hspace{0pt}}m{#1}}

\newcolumntype{Y}{>{\centering\arraybackslash}X}
\newcolumntype{s}{>{\hsize=.25\hsize}Y}
\newcolumntype{t}{>{\hsize=.5\hsize}Y}

\newcommand{\etal}{\mbox{\emph{et al.}}}
\newcommand{\ie}{\mbox{\emph{i. e.}}}

\DeclareMathOperator*{\argmax}{argmax}

\FGfinalcopy 

\IEEEoverridecommandlockouts                              
\title{\LARGE \bf Pose for Action -- Action for Pose}

\author{\vspace{-1mm}Umar Iqbal}
\author{\vspace{-1mm}Martin Garbade}
\author{\vspace{-1mm}Juergen Gall}
\affil{Computer Vision Group, University of Bonn, Germany \\ \texttt{\footnotesize\{uiqbal, garbade, gall\}@iai.uni-bonn.de}\vspace{-3mm}}

\begin{document}

\maketitle

\begin{abstract}

In this work we propose to utilize information about human actions to improve pose estimation in monocular videos. To this end, we present a  pictorial structure model that exploits high-level information about activities to incorporate higher-order part dependencies by modeling action specific appearance models and pose priors. However, instead of using an additional expensive action recognition framework, the action priors are efficiently estimated by our pose estimation framework. This is achieved by starting with a uniform action prior and updating the action prior during pose estimation. We also show that learning the right amount of appearance sharing among action classes improves the pose estimation. We demonstrate the effectiveness of the proposed method on two challenging datasets for pose estimation and action recognition with over 80,000 test images.\footnote{The models and source code are available at \url{http://pages.iai.uni-bonn.de/iqbal_umar/action4pose/}.}

\end{abstract}

\section{Introduction}

Human pose estimation from RGB images or videos is a challenging problem in computer vision, especially for realistic and unconstrained data taken from the Internet. Popular approaches for pose estimation~\cite{desai_eccv2012,pishchulin_cvpr2013, yang_tpami2014,dantone_tpami2014,cherian_cvpr2014,tompson2014joint} adopt the pictorial structure (PS) model, which resembles the human skeleton and allows for efficient inference in case of tree structures \cite{Felzenszwalb_ijcv2005,felzenszwalb_tpami2010}. Even if they are trained discriminatively, PS models struggle to cope with the large variation of human pose and appearance. 
This problem can be addressed by conditioning the PS model on additional observations from the image. 
For instance, \cite{pishchulin_cvpr2013} detects poselets, which are examples of consistent appearance and body part configurations, and condition the PS model on these.

Instead of conditioning the PS model on predicted configurations of body parts from an image, we propose to condition the PS model on high-level information like activity.   
Intuitively, the information about the activity of a person can provide a strong cue about the pose and vice versa the activity can be estimated from pose. There have been only few works~\cite{yao_ijcv2012,Yu_cvpr2013,bruce_cvpr2015} that couple action recognition and pose estimation to improve pose estimation. In~\cite{yao_ijcv2012}, action class confidences are used to initialize an optimization scheme for estimating the parameters of a subject-specific 3D human model in indoor multi-view scenarios. In \cite{Yu_cvpr2013}, a database of 3D poses is used to learn a cross-modality regression forest that predicts the 3D poses from a sequence of 2D poses, which are estimated by~\cite{yang_tpami2014}. In addition, the action is detected and the 3D poses corresponding to the predicted action are used to refine the pose. However, both approaches cannot be applied to unconstrained monocular videos. While \cite{yao_ijcv2012} requires a subject-specific model and several views, \cite{Yu_cvpr2013} requires 3D pose data for training. More recently, \cite{bruce_cvpr2015} proposed an approach to jointly estimate action classes and refine human poses. The approach decomposes the human poses estimated at each video frame into sub-parts and tracks these sub-parts across time according to the parameters learned for each action. The action class and joint locations corresponding to the best part-tracks are selected as estimates for the action class and poses. 
The estimation of activities, however, comes at high computational cost since the videos are pre-processed by several approaches, one for pose estimation \cite{park_iccv2011} and two for extracting action related features~\cite{wang:2011,wang2014cross}. 

\begin{figure*}[t!]
\centering
\captionsetup[figure]{skip=0pt}
\includegraphics[scale=0.935]{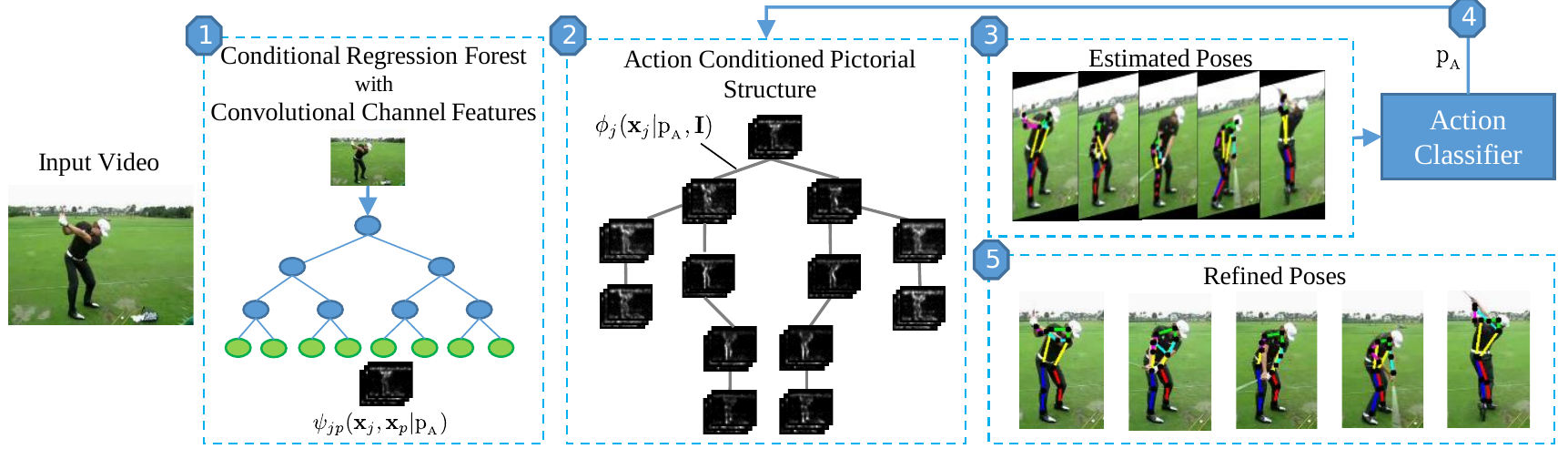}
\caption{Overview of the proposed framework. We propose an action conditioned pictorial structure model for human pose estimation (2). Both the unaries $\phi$ and the binaries $\psi$ of the model are conditioned on the distribution of action classes $\mathrm{p_{_A}}$. While the pairwise terms are modeled by Gaussians conditioned on $\mathrm{p_{_A}}$, the unaries are learned by a regression forest conditioned on $\mathrm{p_{_A}}$ (1). Given an input video, we do not have any prior knowledge about the action and use a uniform prior $\mathrm{p_{_A}}$. We then predict the pose for each frame independently (3). Based on the estimated poses, the probabilities of the action classes $\mathrm{p_{_A}}$ are estimated for the entire video (4). Pose estimation is repeated with the updated action prior $\mathrm{p_{_A}}$ to obtain better pose estimates (5).}
  \vspace{-4mm}
\label{fig:overview}
\end{figure*}

In this work, we present a framework for pose estimation that infers and integrates activities with a very small computational overhead compared to an approach that estimates the pose only. This is achieved by an action conditioned pictorial structure (ACPS) model for 2D human pose estimation that incorporates priors over activities. The framework of the approach is illustrated in Figure~\ref{fig:overview}. We first infer the poses for each frame with a uniform distribution over actions. 
While the binaries of the ACPS are modeled by Gaussian mixture models, which depend on the prior distribution over the action classes, the unaries of the ACPS model are estimated by action conditioned regression forests. To this end, we modify the approach \cite{dantone_tpami2014}, which consists of two layers of random forests, on two counts. Firstly, we replace the first layer by a convolutional network and use convolutional channel features to train the second layer, which consists of regression forests. Secondly, we condition the regression forests on a distribution over actions and learn the sharing among action classes. In our experiments, we show that these modifications increase the pose estimation accuracy by more than 40\% compared to~\cite{dantone_tpami2014}. After the poses are inferred with a uniform distribution over actions, we update the action prior and the ACPS model based on the inferred poses to obtain the final pose estimates. Since the update procedure is very efficient, we avoid the computational expensive overhead of~\cite{bruce_cvpr2015}.

We evaluate our approach on the challenging J-HMDB \cite{Jhuang_iccv2013} and Penn-Action \cite{zhang2013actemes} datasets, which  consist of videos collected from the Internet and contain large amount of scale and appearance variations. In our experiments, we provide a detailed analysis of the impact of conditioning unaries and binaries on a distribution over actions and the benefit of appearance sharing among action classes.  We demonstrate the effectiveness of the proposed approach for pose estimation and action recognition on both datasets. Compared to \cite{bruce_cvpr2015}, the pose estimation accuracy is improved by over 30\%.


\section{Related Work}
State-of-the-art approaches for pose estimation are mostly based on neural networks \cite{toshev2014deeppose, tompson_cvpr2015, Pfister15a, wei2016cvpr, Pishchulin_2016_CVPR, rafi2016bmvc, georgia2016eccv} or on the pictorial structure framework \cite{yang_tpami2014,pishchulin_cvpr2013,dantone_tpami2014, dong2015iccv}.   

Several approaches have been proposed to improve the accuracy of PS models for human pose estimation. For instance, joint dependencies can be modeled not only by the PS model, but also by a mid-level image representation such as poselets~\cite{
pishchulin_cvpr2013}, exemplars~\cite{sap_cvpr2010} or data dependent probabilities learned by a neural network~\cite{chen_nips2014}.  
Pose estimation in videos can be improved by taking temporal information or motion cues into account~\cite{park_iccv2011, cherian_cvpr2014, zuffi_iccv2013, georgia2016eccv, newell2016stacked, insafutdinov16ariv}. In \cite{park_iccv2011} several pose hypotheses are generated for each video frame and a smooth configuration of poses over time is selected from all hypotheses. Instead of complete articulated pose, \cite{ramakrishna_cvpr2013} and \cite{cherian_cvpr2014} track individual body parts and regularize the trajectories of the body parts through the location of neighboring parts. Similar in spirit, the approach in \cite{dong2015iccv} jointly tracks symmetric body parts in order to better incorporate spatio-temporal constraints, and also to avoid double-counting. Optical flow information has also been used to enhance detected poses at each video frame by analyzing body motion in adjacent frames \cite{fragkiadaki2013pose, zuffi_iccv2013}. 

Recent approaches for human pose estimation use different CNN architectures to directly obtain the heatmaps of body parts \cite{tompson_cvpr2015, Pfister15a, wei2016cvpr, Pishchulin_2016_CVPR, rafi2016bmvc, georgia2016eccv}.  \cite{tompson_cvpr2015, wei2016cvpr, Pishchulin_2016_CVPR} and \cite{rafi2016bmvc} use fully convolutional neural network architectures, where \cite{wei2016cvpr} proposes a multi-staged architecture that sequentially refines the output in each stage. For pose estimation in videos, \cite{Pfister15a} combines the heatmaps of body parts from multiple frames with optical flow information to leverage the temporal information in videos. More recently, \cite{georgia2016eccv} proposes a convolutional recurrent neural network architecture that takes as input a set of video frames and sequentially estimates the body part locations in each frame, while also using the information of estimated body parts in previous frames. 

As done in this work, a few works also combine both PS model and CNNs~\cite{chen_nips2014, tompson2014joint}. In contrast to the approaches that use temporal information in videos for pose refinement, we utilize the detected poses in each video frame to extract high-level information about the activity and use it to refine the poses. 

Action recognition based on 3D human poses has been investigated in many works~\cite{ye2013survey}. With the progress in the area of 2D human pose estimation in recent years, 2D poses have also gained an increased attention for action recognition~\cite{YangWM10, singh_iccv2011action, desai_eccv2012, Jhuang_iccv2013, pishculin_gcpr2014, cheron2015p}. However, utilizing action recognition to aid human pose estimation is not well studied, in particular not in the context of 2D human pose estimation. There are only a few works \cite{yao_ijcv2012,Yu_cvpr2013,ukita2013iterative,Bangpeng_TPAMI2012,bruce_cvpr2015} that showed the benefit of it. The approaches in \cite{yao_ijcv2012,Yu_cvpr2013} rely on strong assumption.   
The approach \cite{yao_ijcv2012} assumes that a person-specific 3D model is given and considers pose estimation in the context of multiple synchronized camera views. The approach \cite{Yu_cvpr2013} focuses on 3D pose estimation from monocular videos and assumes that 3D pose data is available for all actions. 
The approach \cite{ukita2013iterative} adopts a mixture of PS models, and learns a model for each action class. For a given image, each model is weighted by the confidence scores of an additional action recognition system and the pose with the maximum weight is taken. A similar approach is adopted in \cite{Bangpeng_TPAMI2012} to model object-pose relations. These approaches, however, do not scale with the number of action classes since each model needs to be evaluated. 

The closest to our work is the recent approach of \cite{bruce_cvpr2015} that jointly estimates the action classes and refines human poses. The approach first estimates human poses at each video frame and decomposes them into sub-parts. These sub-parts are then tracked across video frames based on action specific spatio-temporal constraints. Finally, the action labels and joint locations are inferred from the part tracks that maximize a defined objective function. While the approach shows promising results, it does not re-estimate the parts but only re-combines them over frames \ie , only the temporal constraints are influenced by an activity. Moreover, it relies on two additional activity recognition approaches based on optical flow and appearance features to obtain good action recognition accuracy that results in a very large computational overhead as compared to an approach that estimates activities using only the pose information. In this work, we show that additional action recognition approaches are not required, but predict the activities directly from a sequence of poses. In contrast to \cite{bruce_cvpr2015}, we condition the pose model itself on activities and re-estimate the entire pose per frame.

\section{Overview}\label{sec:overview}

Our method exploits the fact that the information about the activity of a subject provides a cue about pose and appearance of the subject, and vice versa. 
In this work we utilize the high-level information about a person's activity to leverage the performance of pose estimation, where the activity information is obtained from previously inferred poses. To this end, we propose an action conditioned pictorial structure (PS) that incorporates action specific appearance and kinematic models. If we have only a uniform prior over the action classes, the model is a standard PS model, which we will briefly discuss in Section~\ref{sec:PS_model}. Figure~\ref{fig:overview} depicts an overview of the proposed framework.    

\section{Pictorial Structure}\label{sec:PS_model}

We adopt the joint representation~\cite{dantone_tpami2014} of the PS model \cite{Felzenszwalb_ijcv2005}, where the vector $\bold{x}_j \in \mathcal{X}$ represents the 2D location of the $j^{th}$ joint in image $\bold{I}$, and $\mathcal{X}$ = $\{\bold{x}_j\}_{j \in \mathcal{J}}$ is the set of all body joints. The structure of a human body is represented by a kinematic tree with nodes of the tree being the joints ${j}$ and edges $\mathcal{E}$ being the kinematic constraints between a joint $j$ and its unique parent joint $p$ as illustrated in Figure~\ref{fig:overview}. The pose configuration in a single image is then inferred by maximizing the following posterior distribution:
%
%
%
\begin{equation}
\label{eqt:ps}
\mathrm{p}(\mathcal{X}|\bold{I}) \propto \prod_{j \in \mathcal{J}} \phi_j(\bold{x}_j|\bold{I})  \prod_{j,p \in \mathcal{E}} \psi_{jp}(\bold{x}_j, \bold{x}_p)  
\end{equation}
where the unary potentials $\phi_j(\bold{x}_j|\bold{I})$ represent the likelihood of the $j^{th}$ joint at location $\bold{x}_j$. The binary potentials $\psi_{jp}(\bold{x}_j, \bold{x}_p)$ define the deformation cost for the joint-parent pair $(j, p)$, and are often modeled by Gaussian distributions for an exact and efficient solution using a distance transform \cite{Felzenszwalb_ijcv2005}. 
We describe the unary and binary terms in Section \ref{sec:unary_potentials} and Section \ref{sec:binary_potentials}, respectively. In Section~\ref{sec:action_cond_ps}, we then discuss how these can be adapted to build an action conditioned PS model.

\subsection{Unary Potentials}
\label{sec:unary_potentials}

Random regression forests have been proven to be robust for the task of human pose estimation \cite{shotton2011real, minsun_cvpr2012, dantone_tpami2014}. A regression forest $\mathcal{F}$ consists of a set of randomized regression trees, where each tree $T$ is composed of split and leaf nodes. Each split node represents a weak classifier which passes an input image patch $P$ to a subsequent left or right node until a leaf node $L_T$ is reached. As in \cite{dantone_tpami2014}, we use a separate regression forest for each body joint. Each tree is trained with a set of randomly sampled images from the training data. The patches around the annotated joint locations are considered as foreground and all others as background. Each patch consists of a joint label $c \in \{0,j\}$, a set of image features $F_P$, and its 2D offset $\bold{d}_P$ from the joint center. 
During training, a splitting function is learned for each split node by randomly selecting and maximizing a goodness measure for regression or classification.  
At the leaves the class probabilities $\mathrm{p}(c|L_T)$ and the distribution of offset vectors $\mathrm{p}(\bold{d}|L_T)$ are stored.          

During testing, patches are densely extracted from the input image $\bold{I}$ and are passed through the trained trees. 
Each patch centered at location $\bold{y}$ ends in a leaf node $L_T (P(\bold{y}))$ for each tree $T \in \mathcal{F}$.
The unary potentials $\phi_j$ for the joint $j$ at location $\bold{x}_j$ are then given by 
\begin{multline} 
\label{eqt:unary_potentials}
\phi_j(\bold{x}_j|\bold{I}) = \sum_{\bold{y} \in \Omega}\dfrac{1}{|\mathcal{F}|} \sum_{T \in \mathcal{F}} \Big\{ \mathrm {p}(c=j|L_T (P(\bold{y}))) \\ \cdot \mathrm{p}(\bold{x}-\bold{y}|L_T (P(\bold{y}))\Big\}.
\end{multline}

In \cite{dantone_tpami2014} a two layer approach is proposed. The first layer consists of classification forests that classify image patches according to the body parts using a combination of color features, HOG features, and the output of a skin color detector. The second layer consists of regression forests that predict the joint locations using the features of the first layer and the output of the first layer as features. For both layers, the split nodes compare feature values at different pixel locations within a patch of size $24\times24$ pixels.   

\begin{figure}
\centering
\captionsetup[figure]{skip=0pt}
\includegraphics[scale=0.95]{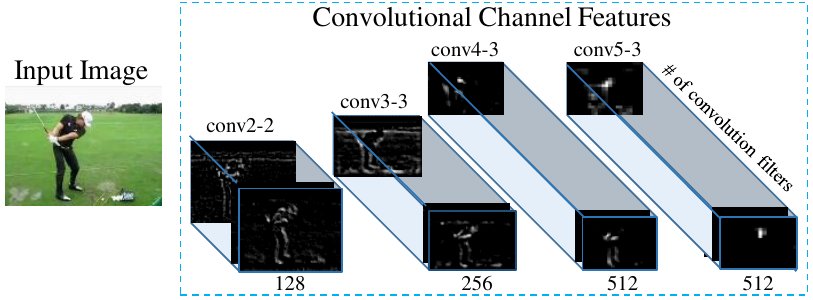}
\caption{Example of convolutional channel features extracted using VGG-16 net \cite{Simonyan14c}.}
\label{fig:ccf}
\end{figure}

We propose to replace the first layer by a convolutional network and extract convolutional channel features (CCF)~\cite{yang_iccv2015} from the intermediate layers of the network to train the regression forests of the second layer. In~\cite{yang_iccv2015} several pre-trained network architectures have been evaluated for pedestrian detection using boosting as classifier. The study shows that the ``conv3-3'' layer of the VGG-16 net~\cite{Simonyan14c} trained on the ImageNet (ILSVRC-2012) dataset performs very well even without fine tuning, but it is indicated that the optimal layer depends on the task. Instead of pre-selecting a layer, we use regression forests to select the features based on the layers ``conv2-2", ``conv3-3", ``conv4-3", and ``conv5-3". An example of the CCF extracted from an image is shown in Figure~\ref{fig:ccf}. Since these layers are of lower dimensions than the original image, we upsample them using linear interpolation to make their dimensions equivalent to the input image. This results in a 1408 (128+256+512+512) dimensional feature representation for each pixel. As split nodes in the regression forests, we use axis-aligned split functions. For an efficient feature extraction at multiple image scales, we use patchwork as proposed in \cite{iandola2014densenet} to perform the forward pass of the convolutional network only once.

\subsection{Binary Potentials}
\label{sec:binary_potentials}
Binary potentials $\psi_{jp}(\bold{x}_j, \bold{x}_p)$ are modeled as a Gaussian mixture model for each joint $j$ with respect to its parent joint $p$ in the kinematic tree. As in \cite{dantone_tpami2014}, we obtain the relative offsets between child and parent joints from the training data and cluster them into $k=1, \dots, K$ clusters using k-means clustering. Each cluster $k$ takes the form of a weighted Gaussian distribution as
\begin{multline}\label{eq:binary}
\psi_{jp}(\bold{x}_j, \bold{x}_p) = w_{jp}^{k} \cdot \\ \exp\left(-\dfrac{1}{2}\left(\bold{d}_{jp}-\mu_{jp}^{k}\right)^T\left(\Sigma_{jp}^{k}\right)^{-1}\left(\bold{d}_{jp}-\mu_{jp}^{k})\right)\right)
\end{multline}
with mean $\mu_{jp}^{k}$ and covariance $\Sigma_{jp}^{k}$, where $\bold{d}_{jp} = (\bold{x}_j-\bold{x}_p)$. The weights $w_{jp}^{k}$ are set according to the cluster frequency $\mathrm{p}(k|j,p)^\alpha$ with a normalization constant $\alpha=0.1$ \cite{dantone_tpami2014}. 

For inference, we select the best cluster $k$ for each joint 
by computing the max-marginals for the root node and backtrack the best pose configuration from the maximum of the max-marginals.  

\section{Action Conditioned Pose Estimation}

As illustrated in Figure~\ref{fig:overview}, our goal is to estimate the pose $\mathcal{X}$ conditioned by the distribution $\mathrm{p_{_A}}$ for a set of action classes $a \in \mathcal{A}$. To this end, we introduce in Section~\ref{sec:action_cond_ps} a pictorial structure model that is conditioned on $\mathrm{p_{_A}}$. Since we do not assume any prior knowledge of the action, we estimate the pose first with the uniform distribution $\mathrm{p_{_A}}(a) = 1/\vert \mathcal{A} \vert, \forall a \in \mathcal{A} $. The estimated poses for $N$ frames are then used to estimate the probabilities of the action classes $\mathrm{p_{_A}}(a|\mathcal{X}_{n=1 \dots N}),  \forall a \in \mathcal{A} $ as described in Section~\ref{sec:action_recognition}. Finally, the poses $\mathcal{X}_n$ are updated based on the distribution $\mathrm{p_{_A}}$.   

\subsection{Action Conditioned Pictorial Structure}
\label{sec:action_cond_ps}

In order to integrate the distribution $\mathrm{p_{_A}}$ of the action classes obtained from the action classifier into \eqref{eqt:ps}, we make the unaries and binaries dependent on $\mathrm{p_{_A}}$: 
\begin{equation}\label{eq:ps_cond}
\mathrm{p}(\mathcal{X}|\mathrm{p_{_A}},\bold{I}) \propto \prod_{j \in \mathcal{J}} \phi_j(\bold{x}_j|\mathrm{p_{_A}},\bold{I}) \cdot \prod_{j,p \in \mathcal{E}} \psi_{jp}(\bold{x}_j, \bold{x}_p|\mathrm{p_{_A}}). 
\end{equation} 
While the unary terms are discussed in Section~\ref{sec:cond_unary_potentials}, the binaries $\psi_{jp}(\bold{x}_j, \bold{x}_p|\mathrm{p_{_A}})$ are represented by Gaussians as in \eqref{eq:binary}. However, instead of computing mean and covariance from all training poses with equal weights, we weight each training pose based on its action class label and $\mathrm{p_{_A}}(a)$. In our experiments, we will also investigate the case where $\mathrm{p_{_A}}(a)$ is simplified to   
 \begin{equation}
 \label{eqt:best_ps}
\mathrm{p_{_A}}(a) = \begin{cases}
         1 & \text{if} \quad a=\argmax_{a'}\mathrm{p_{_A}}(a'|\mathcal{X}_{n=1 \dots N})\\
         0 & \text{otherwise.}
        \end{cases}
\end{equation}

\subsubsection{Conditional Joint Regressors}\label{sec:cond_unary_potentials}
\begin{figure}[t!]
\includegraphics[scale=0.8,trim={0 0 0 0},clip]{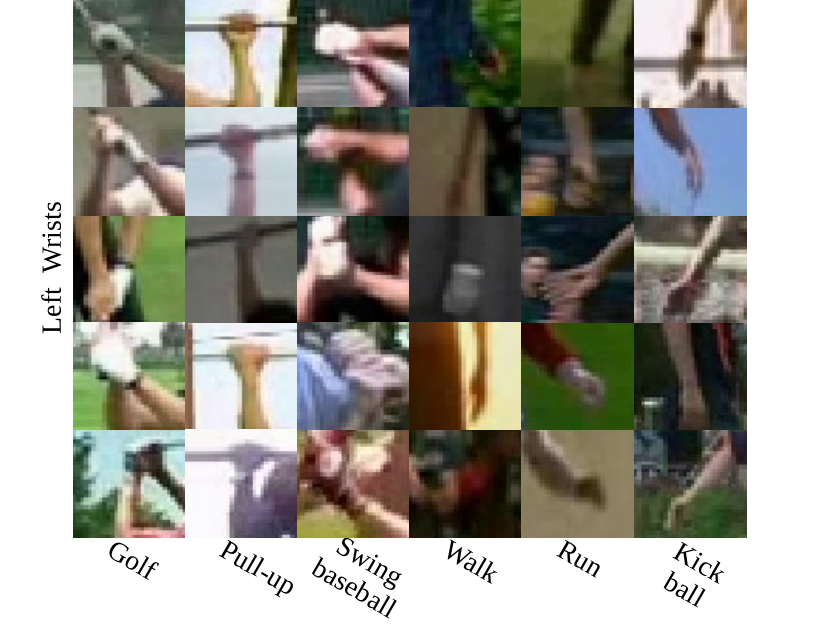}
\centering
\captionsetup[figure]{skip=0pt}
\caption{Example patches centered at the wrist of the left hand side. We can see a large amount of appearance variation for a single body part. However, for several activities, in particular sports such as \textit{golf} and \textit{pull-up}, this variation is relatively small within the action classes. Nonetheless, a few classes also share appearance with each other e.g., \textit{golf} and \textit{baseball} or activities such as \textit{run} and \textit{kick ball}. This clearly shows the importance of class specific appearance models with a right amount of appearance sharing across action classes for efficient human pose estimation.}
\label{fig:appearance_sharing}
\end{figure}

Figure~\ref{fig:appearance_sharing} shows examples of patches of the wrist extracted from images of different action classes. We can see a large amount of appearance variation across different classes regardless of the fact that all patches belong to the same body joint. However, it can also be seen that within individual activities this variation is relatively small. We exploit this observation and propose action specific unary potentials for each joint $j$. To this end we adopt conditional regression forests \cite{dantone_cvpr2012, minsun_cvpr2012} that have been proven to be robust for facial landmark detection in \cite{dantone_cvpr2012} and 3D human pose estimation in \cite{minsun_cvpr2012}. 
While \cite{dantone_cvpr2012} trains a separate regression forest for each head pose and selects a specific regression forest conditioned on the output of a face pose detector, \cite{minsun_cvpr2012} proposes partially conditioned regression forests, where a forest is jointly trained for a set of discrete states of a human attribute like human orientation or height and the conditioning only happens at the leaf nodes. Since the continuous attributes are discretized, interpolation between the discrete states is achieved by sharing the votes.

In this work we resort to partially conditional forests due to its significantly reduced training time and memory requirements. 
During training we augment each patch $P$ with its action class label $a$. Instead of $\mathrm{p}(c|L_T)$ and $\mathrm{p}(\bold{d}|L_T)$, the leaf nodes model the conditional probabilities  $\mathrm{p}(c|a,L_T)$ and $\mathrm{p}(\bold{d}|a,L_T)$. 
Given the distribution over action classes $\mathrm{p_{_A}}$, we obtain the conditional unary potentials:
\begin{align}
\label{eqt:conditional_unary_potentials} 
\nonumber\phi_j(\bold{x}_j|\mathrm{p_{_A}},\bold{I}) &= \sum_{\bold{y} \in \Omega}\dfrac{1}{|\mathcal{F}|} \sum_{T \in \mathcal{F}} \sum_{a \in \mathcal{A}} \Big\{\mathrm{p_{_A}}(a) \\
\nonumber&\hspace{-5mm}\cdot \mathrm{p}(c=j|a,L_T (P(\bold{y}))) . \mathrm{p}(\bold{x}-\bold{y}|a,L_T (P(\bold{y})) \Big\} \\
&= \sum_{a \in \mathcal{A}} \phi_j(\bold{x}_j|a,\bold{I}) \mathrm{p_{_A}}(a).
\end{align}
Since the terms $\phi_j(\bold{x}_j|a,\bold{I})$ need to be computed only once for an image $\bold{I}$, $\phi_j(\bold{x}_j|\mathrm{p_{_A}},\bold{I})$ can be efficiently computed after an update of $\mathrm{p_{_A}}$.  

\subsubsection{Appearance Sharing Across Actions}\label{sec:sharing}
\label{sec:appearance_sharing}

Different actions sometimes share body pose configurations and appearance of parts as shown in Figure~\ref{fig:appearance_sharing}. We therefore propose to learn the sharing among action classes within a conditional regression forest. To this end, we replace the term $\phi_j(\bold{x}_j|a,\bold{I})$ in \eqref{eqt:conditional_unary_potentials} by a weighted combination of the action classes:

\begin{equation}
\label{eqt:unary_with_cond}
\phi^{\textit{sharing}}_j(\bold{x}_j|a, \bold{I}) = \sum_{a'  \in \mathcal{A}} \gamma_{a}(a') \phi_j(\bold{x}_j|a',\bold{I})
\end{equation} 
where the weights $\gamma_{a}(a')$ represent the amount of sharing between action class $a$ and $a'$. To learn the weights $\gamma_{a}$ for each class $a \in \mathcal{A}$, we apply the trained conditional regression forests to a set of validation images scaled to a constant body size and maximize the response of \eqref{eqt:unary_with_cond} at the true joint locations and minimize it at non-joint locations. Formally, this can be stated as
\begin{multline}
\label{eqt:sharing_cost_fun}
\gamma_a = \argmax_{\gamma} \sum_{n_a} \sum_{j} \Bigg\{ \sum_{a' \in \mathcal{A}} \gamma(a') \phi^*_{j}\left(\bold{x}^{\textit{gt}}_{j,n_a}\vert{a'},\bold{I}_{n_a}\right)  \\ - \max_{\bold{x} \in \bold{X}_{j,n_a}^{\textit{neg}}}\left(\sum_{a' \in \mathcal{A}} \gamma(a') \phi^*_{j}\left(\bold{x}\vert{a'},\bold{I}_{n_a}\right)\right) \Bigg\} - \lambda \Vert \gamma \Vert^2
\end{multline}
subject to $\sum_{a'  \in \mathcal{A}} \gamma(a') = 1$ and $\gamma(a') \geq 0$. $\bold{I}_{n_a}$ denotes the $n^{th}$ scaled validation image of action class $a$, $\bold{x}^{\textit{gt}}_{j,n_a}$ is the annotated joint position for joint $j$ in image $\bold{I}_{n_a}$, and $\bold{X}_{j,n_a}^{\textit{neg}}$ is a set of image locations which are more than $5$ pixels away from $\bold{x}^{\textit{gt}}_{j,n_a}$. 
The set of negative samples is obtained by computing $\phi^*_{j}(\bold{x}\vert{a'},\bold{I}_{n_a})$ and taking the $10$ strongest modes, which do not correspond to $\bold{x}^{\textit{gt}}_{j,n_a}$, for each image. For optimization, we use the smoothed unaries
\begin{equation}
\phi_{j}^{*}(\bold{x}|a,\bold{I}) = \sum_{\bold{y} \in \Omega} \exp\left(-\dfrac{{\Vert \bold{x}-\bold{y}\Vert}^2}{\sigma^2}\right) \phi_{j}(\bold{y}|a,\bold{I}) 
\end{equation}
with $\sigma = 3$ and replace $\max$ by the softmax function to make \eqref{eqt:sharing_cost_fun} differentiable. The last term in \eqref{eqt:sharing_cost_fun} is a regularizer that prefers sharing, \ie, $\Vert \gamma \Vert^2$ attains its minimum value for uniform weights. In our experiments, we use $\lambda=0.4$ as weight for the regularizer.    
We optimize the objective function by constrained local optimization using uniform weights for initialization $\gamma(a')=1/\vert\mathcal{A} \vert$. In our experiments, we observed that similar weights are obtained when the optimization is initialized by $\gamma(a)=1$ and $\gamma(a')=0$ for $a'\neq a$, indicating that the results are not sensitive to the initialization.   
In \eqref{eqt:sharing_cost_fun}, we learn the weights $\gamma_a$ for each action class but we could also optimize for each joint independently. In our experiments, however, we observed that this resulted in over-fitting. 

\subsection{Action Classification}
\label{sec:action_recognition}

For pose-based action recognition, we use the bag-of-word approach proposed in~\cite{Jhuang_iccv2013}. 
From the estimated joint positions $\mathcal{X}_{n=1 \dots N}$, a set of features called \emph{NTraj+} is computed that encodes spatial and directional joint information. Additionally, differences between successive frames are computed to encode the dynamics of the joint movements. Since we use a body model with 13 joints, we compute the locations of missing joints (neck and belly) in order to obtain the same $15$ joints as in~\cite{Jhuang_iccv2013}. We approximate the neck position as the mean of the face and the center of shoulder joints. The belly position is approximated by the mean of the shoulder and hip joints. \\
\indent For each of the $3,223$ descriptor types, a codebook is generated by running k-means $8$ times on all training samples and choosing the codebook with minimum compactness. These codebooks are used to extract a histogram for each descriptor type and video. 
For classification, we use an SVM classifier in a multi-channel setup. To this end, for each descriptor type $t$, we compute a distance matrix $D_t$ that contains the $\chi^{2}$-distance between the histograms $(h_{i}^{t},h_{j}^{t})$ of all video pairs $(v_i, v_j)$. We then obtain the kernel matrix that we use for the multi-class classification as follows 
\begin{equation}
K(v_i,v_j) = \exp \left( - \frac{1}{L} \sum^{L}_{t=1} \frac{D_t(h_{i}^{t},h_{j}^{t})}{\mu^{t}} \right)
\end{equation}
where $L$ is the number of descriptor types and $\mu^{t}$ is the mean of the distance matrix $D_t$. For classification we use a one-vs-all approach with $C = 100$ for the SVM.

\section{Experiments}
\label{sec:experiments}
In order to evaluate the proposed method, we follow the same protocol as proposed in \cite{bruce_cvpr2015}. In particular, we evaluate the proposed method on two challenging datasets, namely \mbox{sub-J-HMDB}~\cite{Jhuang_iccv2013}
and the Penn-Action dataset~\cite{zhang2013actemes}. Both datasets provide annotated 2D poses and activity labels for each video. They consist of videos collected from the Internet and contain large amount of scale and appearance variations, low resolution frames, occlusions and foreshortened body poses. This makes them very challenging for human pose estimation. 
While sub-J-HMDB~\cite{Jhuang_iccv2013} comprises videos from 12 action categories with fully visible persons, the Penn-Action dataset comprises videos from 15 action categories with a large amount of body part truncations. As in \cite{bruce_cvpr2015}, we discard the activity class ``playing guitar" since it does not contain any fully visible person. 
For testing on sub-J-HMDB, we follow the 3-fold cross validation protocol proposed by \cite{Jhuang_iccv2013}. On average for each split, this includes $229$ videos for training and $87$ videos for testing with $8,124$ and $3,076$ frames, respectively. 
The Penn-Action dataset consists of $1,212$ videos for training and $1,020$ for testing with $85,325$ and $74,308$ frames, respectively. 
To evaluate the performance of pose estimation, we use the APK (Average Precision of Keypoints) metric \cite{yang_tpami2014,bruce_cvpr2015}. 

\begin{table}[htbp]
  \centering
  \footnotesize
\def\arraystretch{1}

\scalebox{0.9}{
  \begin{tabular}{*{4}{C{3cm} C{3cm}}}
  	\hline 
  	\multicolumn{2}{c}{\textbf{Features}}\\
	HOG, Color, Skin \cite{dantone_tpami2014} &  CCF \\ \hline
   {$36.7$}	&	{$51.5$} \\
	\hline  	    
  \end{tabular}
  }
  \caption{Comparison of the features used in \cite{dantone_tpami2014} with the proposed convolutional channel features (CCF).   APK with threshold 0.1 on split-1 of sub-J-HMDB.}
  \label{tab:mcf_vs_ccf}
\end{table}

\begin{table*}[!htbp]
  \centering
  \footnotesize
\def\arraystretch{1}
   \begin{subtable}{\columnwidth}
     \centering
   \scalebox{1}{
\begin{tabularx}{0.95\columnwidth}{lsss}
    \hline
	\diaghead{\theadfont Diag ColumnmnHead} {\textbf{Unary}}{\textbf{Binary}} 	& Indep.  & Cond. \eqref{eqt:best_ps} & Cond. ($\mathrm{p_{_A}}$) \\
	\hline    	    
	Indep.	+ CCF			  	&		{$51.5$}	  &  {$53.8$}	&	{$51.0$}  \\
	Cond. \eqref{eqt:best_ps} + CCF	&		{$48.9$}	  &  {$49.9$}	&	{$48.4$} \\
	Cond. \eqref{eqt:best_ps} + AS + CCF	&		{$53.8$}	  &  {$\mathbf{55.3}$}		& {$52.9$}  \\
	Cond. ($\mathrm{p_{_A}}$) + CCF					&		{$52.3$}	  &  {$53.1$}	 &	{$52.0$}  \\
	Cond. ($\mathrm{p_{_A}}$)  + AS + CCF				&		{$53.4$}	  &  {$\mathbf{55.1}$}	 &	{$52.5$}  \\
	\hline 
  \end{tabularx}
  }
  \caption{}
  \label{tab:frame_work_analysis_a}
  \end{subtable} 
  \begin{subtable}{\columnwidth}
    \centering
  \scalebox{1}{
\begin{tabularx}{0.95\columnwidth}{lsss}
    \hline
	\diaghead{\theadfont Diag ColumnmnHead} {\textbf{Unary}}{\textbf{Binary}} 	& Indep.  & Cond. \eqref{eqt:best_ps} & Cond. ~~~~($\mathrm{p_{_A}}$) \\
	\hline    	    
	Indep.				  	&		{$36.7$}	  &  {$38.5$}	&	{$36.7$}  \\
	Cond. \eqref{eqt:best_ps} 	&		{$29.3$}	  &  {$32.5$}	&	{$29.7$} \\
	Cond. \eqref{eqt:best_ps} + AS	&		{$38.0$}	  &  {$\mathbf{39.6}$}		& {$37.2$}  \\
	Cond. ($\mathrm{p_{_A}}$) 					&		{$37.0$}	  &  {$39.0$}	 &	{$36.8$}  \\
	Cond. ($\mathrm{p_{_A}}$)  + AS					&		{$38.0$}	  &  {$\mathbf{39.5}$}	 &	{$37.3$}  \\
	\hline 
  \end{tabularx}
  }
  \caption{}
  \label{tab:frame_work_analysis_b}
  \end{subtable} 
  \caption{Analysis of the proposed framework under different settings. Cond. (5) denotes if the action class probabilities $\mathrm{p_{_A}}$ are replaced by \eqref{eqt:best_ps}. \textbf{(a)} using CCF features. \textbf{(b)} using features from \cite{dantone_tpami2014}. (APK threshold: 0.1)} 
  \label{tab:frame_work_analysis}
\end{table*}
\subsection{Implementation Details}

For the Penn-Action dataset, we split the training images half and half into a training set and a validation set. Since the dataset sub-J-HMDB is smaller, we create a validation set by mirroring the training images. The training images are scaled such that the mean upper body size is $40$ pixels. Each forest consists of $20$ trees, where $10$ trees are trained on the training and $10$ on the validation set, with a maximum depth of $20$ and a minimum of $20$ patches per leaf. We train each tree with $50,000$ positive and $50,000$ negative patches extracted from $5,000$ randomly selected images and generate $40,000$ split functions at each node. 
For the binary potentials, we use $k=24$ mixtures per part. 

For learning the appearance sharing among action classes (Section \ref{sec:appearance_sharing}) and training the action classifier (Section \ref{sec:action_recognition}), we use the $10$ trees trained on the training set and apply them to the validation set. The action classifier and the sharing are then learned on the validation set.  

For pose estimation, we create an image pyramid and perform inference at each scale independently. We then select the final pose from the scale with the highest posterior \eqref{eq:ps_cond}. In our experiments, we use $4$ scales with scale factor $0.8$. 
The evaluation of 260 trees (20 trees for each of the 13 joints) including feature extraction takes roughly 15 seconds on average.\footnote{Measured on a 3.4GHz Intel processor using only one core with NVidia GeForce GTX 780 GPU. The image size for all videos in sub-J-HMDB is $320\times240$ pixels.} Inference with the PS model for all 4 scales takes around 1 second. The action recognition with feature computation takes only 0.18 seconds per image and it does not increase the time for pose estimation substantially.

\begin{table*}[!htbp]
  \centering
  \footnotesize
  \setlength{\tabcolsep}{2.8pt} 
\def\arraystretch{1}

\scalebox{1}{

\begin{tabularx}{0.98\linewidth}{lsssssssssss}
   \hline
	\multirow{2}{*}{\textbf{Method}} & \multirow{2}{*}{\textbf{Head}} &	\multirow{2}{*}{\textbf{Sho}} & \multirow{2}{*}{\textbf{Elb}} & \multirow{2}{*}{\textbf{Wri}} &	\multirow{2}{*}{\textbf{Hip}} & \multirow{2}{*}{\textbf{Knee}} & \multirow{2}{*}{\textbf{Ank}} & \textbf{Avg} & \textbf{Avg}\\
	&   &   &   &   &   &   &  & thr.=0.2 & thr.=0.1 \\
    \hline    
   	 Cond.\!\eqref{eqt:best_ps} \!{+}\! AS U.\ {\&}\ Cond.\!\eqref{eqt:best_ps} B.\ \!{+}\! CCF\  &  90.3 & 76.9 & 59.3 & 55.0 & 85.9 & 76.4 & 73.0 & 73.8 & 51.6\\
   	 Cond.\! ($\mathrm{p_{_A}}$) \!{+}\! AS U.\ {\&}\ Cond.\!\eqref{eqt:best_ps} B.\ \!{+}\! CCF\ & 90.1 & 76.7 & 59.2 & 54.7 & 85.6 & 76.2 & 72.9 & 73.6 & 51.2 \\   	
	Indep.\ U.\ {\&}\ Indep.\ B.\ \!{+}\! CCF &  88.1 & 76.3 & 57.0 & 49.2 & 85.0 & 75.4 & 71.7 & 71.8 & 48.7\\	
	\hline 
				\multicolumn{10}{c}{\textit{State-of-the-art approaches}} \\
    \hline 
	Indep.\ U.\ {\&}\ Indep.\ B.\ \cite{dantone_tpami2014} & 65.6 & 56.4 & 39.1 & 31.1 & 65.2 & 62.8 & 60.9 & 54.4  & 34.4\\
	Yang \& Ramanan \cite{yang_tpami2014} & 73.8 & 57.5 & 30.7 & 22.1 &  69.9 & 58.2 & 48.9 & 51.6 & \textemdash \\
	Park \& Ramanan \cite{park_iccv2011} & 79.0 & 60.3 & 28.7 & 16.0 & 74.8 & 59.2 & 49.3 & 52.5 & \textemdash \\
	Cherian \etal \cite{cherian_cvpr2014} & 47.4 & 18.2 & 0.08 & 0.07 & \textemdash & \textemdash & \textemdash & 16.4 & \textemdash\\
	Nie \etal \cite{bruce_cvpr2015} & 80.3 & 63.5 & 32.5 & 21.6 & 76.3 & 62.7 & 53.1 & 55.7 & \textemdash \\
 	Chen \& Yuille \cite{chen_nips2014} &  78.7 & 68.4 & 48.3 & 39.7 & 76.3 & 66.3 & 60.3 & 62.6 & {42.2}  \\     
 	\hline
  \end{tabularx}
  }
  \caption{Comparison with the state-of-the-art on sub-J-HMDB using APK threshold $0.2$. In the last column, the average accuracy for the threshold $0.1$ is given.}
  \label{tab:final_acc}
\end{table*}

\begin{table*}[t]
  \centering
  \footnotesize
\def\arraystretch{1}
\setlength{\tabcolsep}{2.8pt} 
\scalebox{1}{

\begin{tabularx}{0.98\linewidth}{lsssssssssss}
    \hline
	\multirow{2}{*}{\textbf{Method}} & \multirow{2}{*}{\textbf{Head}} &	\multirow{2}{*}{\textbf{Sho}} & \multirow{2}{*}{\textbf{Elb}} & \multirow{2}{*}{\textbf{Wri}} &	\multirow{2}{*}{\textbf{Hip}} & \multirow{2}{*}{\textbf{Knee}} & \multirow{2}{*}{\textbf{Ank}} & \textbf{Avg} & \textbf{Avg}\\
	&   &   &   &   &   &   &  & thr.=0.2 & thr.=0.1 \\
    \hline    
   	 Cond.\!\eqref{eqt:best_ps} \!{+}\! AS U.\ {\&}\ Cond.\!\eqref{eqt:best_ps} B.\ \!{+}\! CCF\  & 89.1 & 86.4 & 73.9 & 73.0 & 85.3 & 79.9 & 80.3 & 81.1 & 64.8 \\
   	Indep.\ U.\ {\&}\ Indep.\ B.\ \!{+}\! CCF\ & 84.5 & 81.3 & 66.2 & 62.6 & 82.4 & 75.1 & 76.5 & 75.5  & 57.3\\
	\hline 
				\multicolumn{10}{c}{\textit{State-of-the-art approaches}} \\
    \hline 
	Yang \& Ramanan \cite{yang_tpami2014} & 57.9 & 51.3 & 30.1 & 21.4 & 52.6 & 49.7 & 46.2 & 44.2 & \textemdash \\
	Park \& Ramanan \cite{park_iccv2011} & 62.8 & 52.0 & 32.3 & 23.3 & 53.3 & 50.2 & 43.0 & 45.3 & \textemdash \\
	Nie \etal \cite{bruce_cvpr2015} & 64.2 & 55.4 & 33.8 & 24.4 & 56.4 & 54.1 & 48.0 & 48.0 & \textemdash \\
	Gkioxari \etal \cite{georgia2016eccv} & 95.6 & 93.8 & 90.4 & 90.7 & 91.8 & 90.8 & 91.5 & 91.8 & \textemdash \\
	\hline
  \end{tabularx}
  }
  \caption{Comparison with the state-of-the-art in terms of joint localization error on the Penn-Action dataset.} 
  \label{tab:final_acc_penn}
\end{table*}

\subsection{Pose Estimation}

We first evaluate the impact of the convolutional channel features (CCF) for pose estimation on split-1 of sub-J-HMDB. The results in Table \ref{tab:mcf_vs_ccf} show that the CCF outperform the combination of color features, HOG features, and the output of a skin color detector, which is used in \cite{dantone_tpami2014}.

In Table \ref{tab:frame_work_analysis_a} we evaluate the proposed ACPS model under different settings on split-1 of sub-J-HMDB when using CCF features for joint regressors. We start with the first step of our framework where neither the unaries nor the binaries depend on the action classes. This is equivalent to the standard PS model described in Section~\ref{sec:PS_model}, which achieves an average joint estimation accuracy of $51.5\%$. Given the estimated poses, the pose-based action recognition approach described in Section~\ref{sec:action_recognition} achieves an action recognition accuracy of $66.3\%$ for split-1.

Having estimated the action priors $\mathrm{p_{_A}}$, we first evaluate action conditioned binary potentials while keeping the unary potentials as in the standard PS model. As described in Section~\ref{sec:action_cond_ps}, we can use in our model the probabilities $\mathrm{p_{_A}}$ or replace them by the distribution \eqref{eqt:best_ps}, which considers only the classified action class. The first setting is denoted by ``Cond.\ ($\mathrm{p_{_A}}$)'' and the second by ``Cond.\ \eqref{eqt:best_ps}''. It can be seen that the conditional binaries based on \eqref{eqt:best_ps} already outperform the baseline by improving the accuracy from $51.5\%$ to $53.8\%$. However, taking the priors from all classes slightly decreases the performance. This shows that conditioning the binary potentials on the most probable class is a better choice than using priors from all classes. 

Secondly, we analyze how action conditioned unary potentials affect pose estimation. For the unaries, we have the same options ``Cond.\ ($\mathrm{p_{_A}}$)'' and ``Cond.\ \eqref{eqt:best_ps}'' as for the binaries. In addition, we can use appearance sharing as described in Section~\ref{sec:appearance_sharing}, which is denoted by ``AS''. For all three binaries, the conditional unaries based on \eqref{eqt:best_ps} decrease the performance.
Since the conditional unaries based on \eqref{eqt:best_ps} are specifically designed for each action class, they do not generalize well in case of a misclassified action class.. 
However, adding appearance sharing to the conditional unaries boost the performance for both conditioned on \eqref{eqt:best_ps} and $\mathrm{p_{_A}}$. Adding appearance sharing outperforms  all other unaries without appearance sharing, \ie, conditional unaries, independent unaries and the unaries conditioned on $\mathrm{p_{_A}}$. For all unaries, the binaries conditioned on \eqref{eqt:best_ps} outperform the other binaries.    
This shows that appearance sharing and binaries conditioned on the most probable class performs best, which gives an improvement of the baseline from $51.5\%$ to $55.3\%$. 

In Table \ref{tab:frame_work_analysis_b}, we also evaluate the proposed ACPS when using the weaker features from \cite{dantone_tpami2014}. Although the accuracies as compared to CCF features are lower, the benefit of the proposed method remains the same. For the rest of this paper, we will use CCF for all our experiments.

In Table \ref{tab:final_acc} we compare the proposed action conditioned PS model with other state-of-the-art approaches on all three splits of sub-J-HMDB. In particular, we provide a comparison with \cite{dantone_tpami2014, yang_tpami2014, bruce_cvpr2015, park_iccv2011, cherian_cvpr2014, chen_nips2014}. The accuracies for the approaches~\cite{yang_tpami2014, bruce_cvpr2015, park_iccv2011, cherian_cvpr2014} are taken from~\cite{bruce_cvpr2015} where the APK threshold $0.2$ is used. We also evaluated the convolutional network based approach \cite{chen_nips2014} using the publicly available source code trained on sub-J-HMDB. Our approach outperfroms the other methods by a margin, and notably improves wrist localization by more than 5\% as compared to the baseline. 

Table~\ref{tab:final_acc_penn} compares the proposed ACPS with the state-of-the-art on the Penn-Action dataset. The accuracies for the approaches~\cite{yang_tpami2014, bruce_cvpr2015, park_iccv2011} are taken from~\cite{bruce_cvpr2015}. We can see that the proposed method improves the baseline from 75.5\% to 81.1\%, while improving the elbow and wrist localization accuracy by more than 7\% and 10\%, respectively.  The proposed method also significantly outperforms other approaches Only the approach \cite{georgia2016eccv} achieves a higher accuracy than our method. \cite{georgia2016eccv}, however, uses a better multi-staged CNN architecture as baseline compared to our network for computing CCF features. Since the gain of ACPS compared to our baseline even increases when better features are used as shown in  Table~\ref{tab:frame_work_analysis_a} \& Table~\ref{tab:frame_work_analysis_b}, we expect at least a similar performance gain when we use the baseline architecture from \cite{georgia2016eccv} for ACPS.

\begin{table*}[t]
		\footnotesize 
		\begin{center}
			\setlength{\tabcolsep}{1pt}
			\def\arraystretch{1}
			\scalebox{1}{
			\begin{tabular}{lC{3.5cm}C{2.5cm}C{2.5cm}C{2.5cm}C{2.5cm}}				
			\cline{2-6}
			\multicolumn{1}{l}{}     & {Indep. U. + Indep. B. + CCF}	& \multicolumn{4}{c}{Cond. \eqref{eqt:best_ps}+AS U. \& Cond.  \eqref{eqt:best_ps} B. + CCF}	\\		\cline{3-6}			
			\multicolumn{1}{l}{}	& {} & Pose  & IDT-FV ~\cite{wang2013action} & {Pose+IDT-FV} & {GT}\\	 \hline 	
			sub-J-HMDB (split-1)~~~~~~~	&		 { $51.5$ } 	  	& { $55.3~(56.2\%)$ } 	& { $52.6~(66.3\%)$ }      	& { $55.3~(76.4\%)$ }  	& { $55.9~(100\%)$ }	\\
			Penn-Action	&	{ $57.3$ } & { $64.8~(79.0\%)$ } &  \textemdash & \textemdash &  { $68.1~(100\%)$ } \\ \hline						 
				\end{tabular}
				}
			\end{center}
			\vspace{-3mm}
			\caption{Analysis of pose estimation accuracy with respect to action recognition accuracy. The values in the parentheses are the corresponding action recognition accuracies. (APK threshold: 0.1) 
		}
		\captionsetup[figure]{skip=pt}
		\label{tab:class_wise_acc}
\end{table*}

\subsection{Action Recognition} In Table \ref{tab:AR_JHMDB}, we compare the action recognition accuracy obtained by our approach with state-of-the-approaches for action recognition. On sub-J-HMDB, the obtained accuracy using only pose as feature is comparable to the other approaches. Only the recent work \cite{cheron2015p} which combines pose, CNN, and motion features achieves a better action recognition accuracy. However, if we combine our pose-based action recognition with Fisher vector encoding of improved dense trajectories~\cite{wang2013action} using late fusion, we outperform other methods that also combine pose and appearance. The results are similar on the Penn-Action dataset.

In Table \ref{tab:class_wise_acc}, we report the effect of different action recognition approaches on pose estimation. We report the pose estimation accuracy for split-1 of sub-J-HMDB when the action classes are not inferred by our framework, but estimated using improved dense trajectories with Fisher vector encoding (IDT-FV)~\cite{wang2013action} or the fusion of our pose-based method and IDT-FV. Although the action recognition rate is higher when pose and IDT-FV are combined, the pose estimation accuracy is not improved. If the action classes are not predicted but are provided (GT),
the accuracy improves slightly for sub-J-HMDB and
from 64.8\% to 68.1\% for the Penn-Action dataset. We also experimented with several iterations in our framework, but the improvements compared to the achieved accuracy of $51.6\%$ were not more than 0.1\% on all three splits of sub-J-HMDB.  

\begin{table}[t]
  \centering
  \footnotesize
\def\arraystretch{1}

\scalebox{1}{
\begin{tabularx}{0.85\columnwidth}{lss}
    \hline
	\textbf{Method} 	 	& 		\textbf{sub-J-HMDB}  &  \textbf{Penn-Action} \\
	\hline    	    
	 \multicolumn{3}{c}{\textit{Appearance features only}} \\
	 \hline
	 Dense~\cite{Jhuang_iccv2013}	& {46.0\%} & {\textemdash}  \\			 
   	 IDT-FV~\cite{wang2013action}	&	{60.9\%}  & {92.0\%}\\ 				\hline 
	 \multicolumn{3}{c}{\textit{Pose features only}} \\	\hline
	Pose~\cite{Jhuang_iccv2013}	&	{54.1\%}   & {\textemdash}\\	 
	Pose (Ours)		&	{61.5\%}  & {79.0\%}\\	\hline 

  	\multicolumn{3}{c}{\textit{Pose + Appearance features}} \\	\hline  MST~\cite{wang2014cross} &  {45.3\%}  & {74.0\%} \\		 			 
	Pose \!{+}\! Dense~\cite{Jhuang_iccv2013}	&	{52.9\%}   & {\textemdash} \\
	AOG~\cite{bruce_cvpr2015} 		&	{61.2\%}  & {85.5\%}\\	
	P-CNN~\cite{cheron2015p} & {66.8\%} & {\textemdash} \\
	Pose (Ours) \!{+}\! IDT-FV  & 74.6\% &  92.9\% \\
	\hline 	
  \end{tabularx}
  }
  \caption{Comparison of action recognition accuracy with the state-of-the-art approaches on sub-J-HMDB and Penn-Action datasets. }
  \label{tab:AR_JHMDB}
\end{table}

\section{Conclusion}

In this paper, we have demonstrated that action recognition can be efficiently utilized to improve human pose estimation on realistic data. To this end, we presented a pictorial structure model that incorporates high-level activity information 
by conditioning the unaries and binaries on a prior distribution over action labels. Although the action priors can be estimated by an accurate, but expensive action recognition system, we have shown that the action priors can also be efficiently estimated during pose estimation without substantially increasing the computational time of the pose estimation. In our experiments, we thoroughly analyzed various combinations of unaries and binaries and showed that learning the right amount of appearance sharing among action classes improves the pose estimation accuracy. While we expect further improvements by using a more sophisticated CNN architecture as baseline and by including a temporal model, the proposed method has already shown its effectiveness on two challenging datasets for pose estimation and action recognition.  \\

\noindent\textbf{Acknowledgements:} The work was partially supported by the ERC Starting Grant ARCA (677650).

\bibliographystyle{abbrv}
\bibliography{pose_bib}

\end{document}